\documentclass[10pt,twocolumn,letterpaper]{article}

\usepackage[]{cvpr}      

\newcommand{\vscro}{\textsc{SCRO-Bench}}

\definecolor{cvprblue}{rgb}{0.21,0.49,0.74}
\usepackage[pagebackref,breaklinks,colorlinks,allcolors=cvprblue]{hyperref}
\usepackage{tabularx}
\usepackage{multirow}
\usepackage{booktabs} 
\usepackage{longtable} 
\usepackage{arydshln}   
\usepackage{ltablex}
\usepackage{xltabular}
\usepackage{xcolor}
\usepackage{supertabular}
\usepackage{xspace}
\definecolor{myred}{HTML}{D62728}
\definecolor{myblue}{HTML}{1F77B4}
\usepackage[most]{tcolorbox}
\usepackage{graphicx}
\usepackage{cuted}
\usepackage{caption}

\def\confName{CVPR}
\def\confYear{2027}

\title{Do MLLMs Have Human-Like Spatial Commonsense?}

\author{
Xiaofan Lin\textsuperscript{1,2},
Shiwei Zhang\textsuperscript{1},
Peng Zhang\textsuperscript{3},
Yanhao Wu\textsuperscript{3},\\
Shaokun Wang\textsuperscript{4},
Dongyang Zhang\textsuperscript{1},
Xinyu Zhou\textsuperscript{1},
Guiduo Duan\textsuperscript{1}\\[4pt]
\textsuperscript{1}University of Electronic Science and Technology of China (uestc.edu.cn)\\
\textsuperscript{2}Guangdong Ocean University (gdou.edu.cn)\\
\textsuperscript{3}Xi'an Jiaotong University (xjtu.edu.cn)\\
\textsuperscript{4}Harbin Institute of Technology, Shenzhen  (hit.edu.cn)
}

\begin{document}
	\maketitle
	\begin{abstract}
Spatial commonsense is an innate human ability that enables people to understand scene layouts and infer unseen content. Despite rapid advances in multimodal large language models (MLLMs), whether they possess human-like spatial commonsense remains an open question. To this end, we introduce SCRO-Bench (Spatial Commonsense Reasoning using Occlusion), a benchmark that uses occlusion to evaluate whether MLLMs possess spatial commonsense and the ability to leverage it in reasoning. The benchmark introduces occlusion into images of object clusters drawn from counting datasets, requiring models to apply spatial commonsense cues (\eg, cluster boundaries and layout completeness) to determine whether a specified occluded region covers at least one instance of the queried category. SCRO-Bench comprises 1,672 natural images, covering 120 object categories, along with two controlled synthetic test sets. We evaluate eight MLLMs and establish a human baseline, finding that the evaluated models perform substantially worse than humans and exhibit higher false positive rates than false negative rates overall. Further diagnostic and commonsense guidance experiments suggest that models possess some spatial commonsense but have yet to attain human-like competence, while explicit commonsense guidance can improve their performance. SCRO-Bench provides a foundation for evaluating and improving the capabilities of MLLMs that utilize spatial commonsense in reasoning. Project page: \url{https://wuxianjingong123.github.io/scro-bench/}
\end{abstract}
	\section{Introduction}
\label{sec:intro}
Spatial commonsense provides an important foundation for humans to understand scene layouts and infer unseen content, helping them predict physical events, anticipate potential risks, and plan everyday actions~\cite{battaglia2013simulation,fischer2016functional,yun2018temporal}. This capability is also relevant to real-world applications of multimodal large language models (MLLMs), \eg, robotics, embodied AI, and autonomous driving~\cite{driess2023palm,brohan2023rt,sima2024drivelm}.

Existing studies evaluate the spatial perception and reasoning capabilities of MLLMs through tasks involving spatial relation judgments, object localization and orientation recognition, relative depth estimation, and multi-view understanding~\cite{fu2024blink,wang2024picture,ma20253dsrbench,wang2025site}. However, these benchmarks primarily focus on specific spatial tasks or overall spatial capabilities, without making spatial commonsense a central evaluation objective. In real-world applications, models need to apply spatial commonsense appropriately to support reliable scene interpretation and action planning. Deficiencies in this capability may lead to incorrect decisions and pose potential safety risks. This raises an important question: how can we design tasks that specifically assess models' spatial commonsense and clearly distinguish them, through their formulation, from basic spatial perception and understanding tasks?

\begin{figure*}[t]
	\centering
	\includegraphics[width=\textwidth]{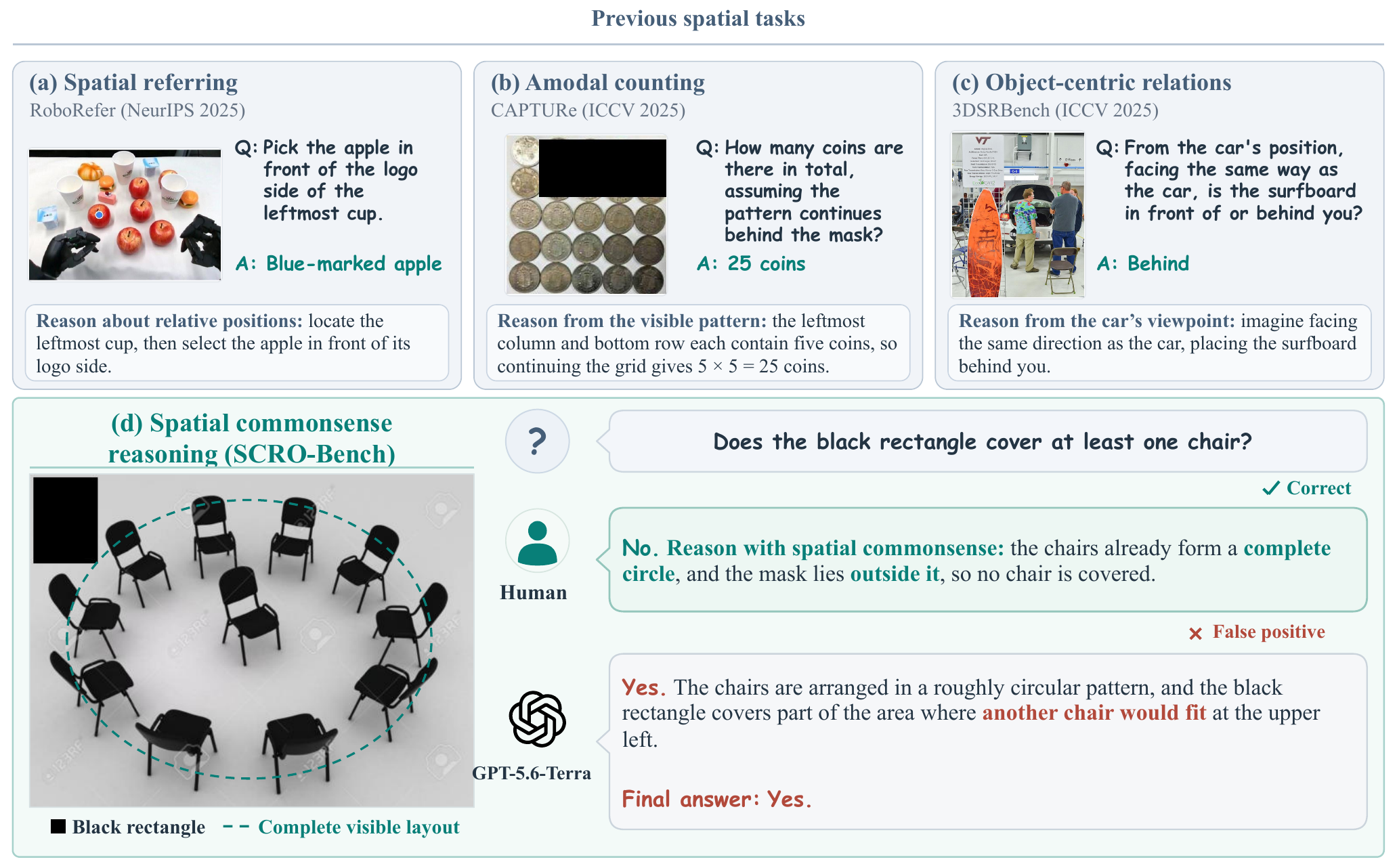}
	\vspace{-2em}
	\caption{Comparison of previous spatial tasks and SCRO-Bench.
		(a) Spatial referring~\cite{zhou2026roborefer}.
		(b) Amodal counting~\cite{pothiraj2025capture}.
		(c) Object-centric relation judgments~\cite{ma20253dsrbench}.
		(d) SCRO-Bench evaluates whether MLLMs can use spatial commonsense to infer target presence in a specified occluded region.
		The mask lies outside a complete circular chair arrangement, yet GPT-5.6-Terra incorrectly answers \texttt{Yes}, producing a false positive.
		The illustrative human rationale uses layout completeness to reach the correct answer.
		The dashed ellipse highlights the complete visible layout.}
	\vspace{-1em}
	\label{fig:task-comparison}
\end{figure*}

To address this, we introduce SCRO-Bench (Spatial Commonsense Reasoning using Occlusion), a benchmark that uses occlusion to evaluate whether models can reason with human-like spatial commonsense. As illustrated in Fig.~\ref{fig:task-comparison}, unlike tasks emphasizing spatial understanding and reasoning (\eg, spatial referring~\cite{zhou2026roborefer}, amodal counting~\cite{pothiraj2025capture}, and object-centric relation judgments~\cite{ma20253dsrbench}), SCRO-Bench makes the appropriate use of spatial commonsense its core evaluation objective. Models are required to use the overall layout of the visible scene to infer whether a specified occluded region covers at least one instance of the queried category. Specifically, we select images containing object clusters from counting datasets, using the visible clusters to establish a scene prior for target presence and applying black masks to occlude local regions. Crucially, this prior may encourage models to overgeneralize target presence from visible regions to occluded regions, inducing a bias toward affirmative predictions. Models therefore need to build on their understanding of visible spatial relations and integrate spatial commonsense cues, including cluster boundaries, layout completeness, and spatial continuity, to determine whether this prior applies to the specified occluded region and reliably infer target presence. Fig.~\ref{fig:scro-overview} provides an overview of the SCRO-Bench task.

We evaluate multiple MLLMs on SCRO-Bench. As shown in Fig.~\ref{fig:task-comparison}, GPT-5.6-Terra incorrectly predicts that the black mask covers a chair even though the mask lies outside the complete circular arrangement, whereas humans can make the correct judgment based on layout completeness. Further evaluations on another seven MLLMs reveal that these models exhibit higher false positive rates than false negative rates overall, suggesting that current MLLMs exhibit some spatial commonsense but have yet to attain human-like proficiency in spatial commonsense reasoning. These challenges reveal limitations in the spatial commonsense reasoning of current MLLMs. Addressing these limitations is critical for MLLMs to operate effectively in real-world. Our main contributions are summarized as follows:

\begin{itemize}
	\item We introduce \vscro{}, a benchmark that makes spatial commonsense its core evaluation objective, comprising a natural-image evaluation set and two controlled synthetic test sets. The natural-image evaluation set contains 1,672 images covering 120 object categories, with ground-truth labels determined from the original unoccluded images, to evaluate models in real-world visual contexts. The synthetic test sets use controlled spatial layouts to support further analysis of factors influencing model judgments.
	
	\item We evaluate eight MLLMs on the natural-image evaluation set and establish a human baseline, finding that the evaluated models underperform humans overall and exhibit substantially higher false positive rates than false negative rates. These results suggest that models may be biased by the target presence prior established by visible object clusters and fail to reliably apply spatial commonsense to constrain their judgments, overgeneralizing target presence from visible to occluded regions.
	
	\item We further investigate model behavior through controlled diagnostic and commonsense guidance experiments. Our findings suggest that although models already possess some spatial commonsense, it remains far from sufficient, as explicit commonsense guidance can significantly improve their performance.
\end{itemize}

	\section{Related Work}
\label{sec:formatting}

\subsection{Visual Spatial Reasoning and Commonsense}
Prior work on visual commonsense reasoning and spatial relation understanding has evaluated models' ability to understand scene structure and reason from context~\cite{liu2023visual,kamath2023s}. VCR~\cite{zellers2019recognition} assesses situational commonsense about actions and intentions through question answering and rationale selection. SPHERE~\cite{zhang2025sphere} provides a multitask evaluation of spatial perception and reasoning, including inference about occluded objects. SCRO-Bench differs from these works in its evaluation focus, emphasizing spatial commonsense rather than situational commonsense about actions and intentions. Moreover, visible instances of the queried category already form clusters in our task, which may encourage models to overgeneralize object presence from visible regions to the masked region. Therefore, understanding spatial commonsense and applying it appropriately are particularly important for our task.

\subsection{Visual Reasoning under Occlusion}
Occlusion limits the visual information available to models, posing additional challenges to visual understanding and reasoning. Prior work~\cite{li2025unveiling,tai2025segment} has therefore sought to evaluate and improve model capabilities using occlusion. CAPTURe~\cite{pothiraj2025capture} evaluates MLLMs' amodal counting ability by adding black masks to regularly arranged objects. O-Bench~\cite{liu2025beyond} constructs occlusion scenarios around target instances to evaluate occlusion relationship identification, complete appearance reasoning, and occlusion rate estimation. In contrast, SCRO-Bench focuses on determining whether a specified mask covers at least one instance of a queried category based on the overall layout of object clusters. Our scenes do not require strictly repeating patterns. Instead, we assess whether models can use spatial commonsense about cluster boundaries, layout completeness, and spatial continuity to infer the presence or absence of target objects.

\subsection{Hallucination in MLLMs}

MLLMs may generate descriptions or judgments inconsistent with image content. Prior work has investigated the evaluation and causes of hallucinations~\cite{rohrbach2018object,guan2024hallusionbench,wu2024evaluating}. POPE~\cite{li2023evaluating} evaluates object hallucination through image-level object existence questions. AMBER~\cite{wang2023amber} further extends the evaluation to existence, attribute, and relation hallucinations. Among these benchmarks, POPE is particularly relevant to SCRO-Bench, as both assess errors in object existence judgments. However, their evaluation targets differ: the former assesses object presence at the image level, whereas the latter examines whether models can combine visible scene layout with spatial commonsense to infer whether a specified mask covers an instance of the queried category. In this setting, even when the queried category is present in the visible region, the mask may cover only background. As shown in Sec.~\ref{sec:results}, the evaluated MLLMs exhibit higher FPRs than FNRs overall. We therefore focus on false positive predictions when the mask covers only background, examining whether models can use spatial commonsense to constrain their inferences about hidden target presence.
	\section{\vscro{}}

\begin{figure*}[t]
    \centering
    \includegraphics[width=\textwidth]{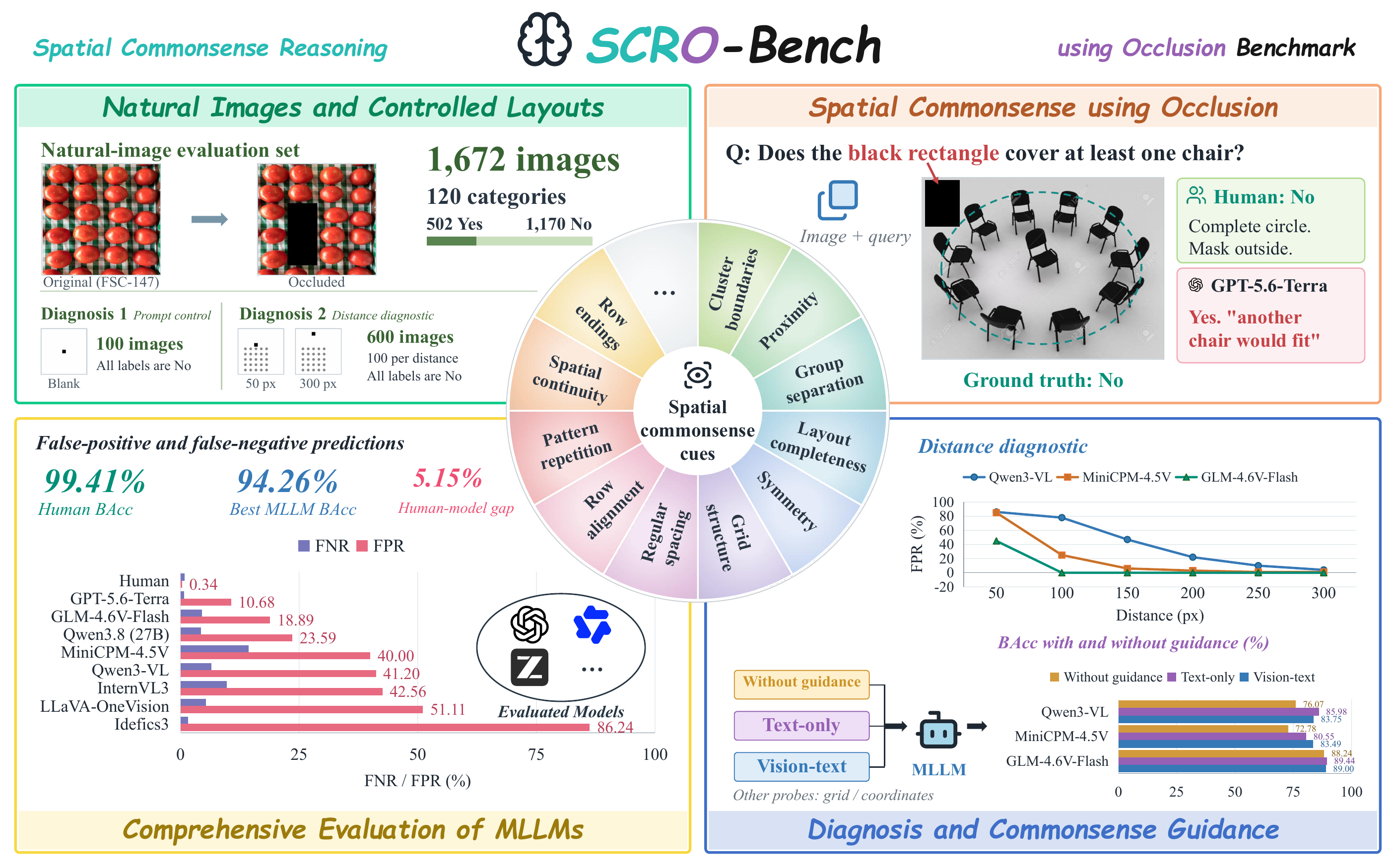}
    \vspace{-2em}
    \caption{\textbf{Overview of SCRO-Bench.}
SCRO-Bench evaluates whether MLLMs can
use spatial commonsense to infer target presence in occluded regions.
We combine natural images with controlled synthetic diagnostics
to assess model performance and examine the effects of commonsense guidance.}
    \vspace{-1em}
    \label{fig:scro-overview}
\end{figure*}
\subsection{Overview}
We introduce \vscro{}, a benchmark that uses occlusion to evaluate whether MLLMs can reason with human-like spatial commonsense. We select images containing object clusters from counting datasets and overlay one black rectangle on each image. Given an occluded image and a queried object category, the model generates a free-form response indicating whether the rectangle covers at least one target instance. An external model then extracts the prediction as \texttt{Yes}, \texttt{No}, or \texttt{Unknown}. Ground-truth labels are determined from the corresponding unoccluded images. Samples are labeled as object-containing if the rectangle overlaps with at least one target instance, and background-only otherwise. Even when the queried category is visible elsewhere, the mask may cover only background. Models therefore need to combine visible layout with spatial commonsense to infer target presence in the masked region. \vscro{} comprises a natural-image evaluation set and two controlled synthetic test sets, supporting evaluation in natural visual contexts and controlled analyses of model behavior.

\subsection{Dataset}
\noindent\textbf{Natural-image evaluation.}
The natural-image set evaluates whether models can infer target presence
in occluded regions of natural scenes. We select images with visible
spatial structures from FSC-147~\cite{ranjan2021learning}, a
few-shot object counting dataset containing 6,135 images from 147 object
categories. As summarized in Tab.~\ref{tab:dataset-statistics}, the
evaluation set contains 1,672 samples across 120 target categories,
including 502 object-containing and 1,170 background-only samples. We deliberately retain more background-only samples to examine whether models can use the spatial commonsense to infer that the black rectangle is more likely to cover background. In these samples, the queried category remains visible elsewhere in the image, so a model cannot correctly answer \texttt{Yes} merely by recognizing the global presence of the category. Object-containing samples prevent a trivial strategy that always answers \texttt{No}.

\noindent\textbf{Controlled synthetic diagnostics.}
We use two controlled synthetic settings to investigate whether the base
prompt contributes to false-positive predictions and whether models exhibit
basic spatial commonsense when reasoning about occluded regions. The first setting is a blank-background prompt control, which tests whether the base prompt itself induces false positive predictions. It contains only a black rectangle overlaid on a blank background, without any instance of the queried category. All samples therefore have the ground-truth label \texttt{No}. We refer to this setting as \textbf{Diagnosis 1}. The second setting is a distance diagnostic designed to probe whether models exhibit basic spatial commonsense when reasoning about occluded regions. We construct a regular $5\times5$ array of circular balls on a plain background. We fix the number, arrangement, and size of the balls, as well as the size and relative direction of the black rectangle. We vary only the distance between the edge of the black rectangle and the ball array. The black rectangle always lies outside the array, and all samples have the ground-truth label \texttt{No}. This setting supports further analysis of the spatial commonsense reflected in model judgments and its limitations. We refer to this setting as \textbf{Diagnosis 2}. Diagnosis~1 contains 100 images, and each distance condition in Diagnosis~2 contains 100 images. Tab.~\ref{tab:dataset-statistics} summarizes the sizes and label
distributions of both diagnostic sets.

\subsection{Statistics and Examples}
Tab.~\ref{tab:dataset-statistics} summarizes the statistics of each
component in \vscro{}, including the number of samples, the number
of queried object categories, and the numbers of object-containing
and background-only samples. Tab.~\ref{tab:benchmark-comparison} compares \vscro{} with previous spatial understanding and reasoning benchmarks
in terms of image counts, object-category coverage, and evaluation tasks. The upper-left panel of Fig.~\ref{fig:scro-overview} shows examples from the natural-image evaluation set and the synthetic test sets. Additional examples are provided in
Supplementary Materials~\ref{app:addition_vis_examples}.



	\section{Experimental Results and Analysis}

\subsection{Experimental Setup}

\paragraph{Models.}
We evaluate \vscro{} using several mainstream MLLMs, including LLaVA-OneVision~\cite{li2024llava}, Idefics3~\cite{laurenccon2024building}, Qwen3-VL~\cite{yang2025qwen3}, MiniCPM-V-4.5~\cite{yu2026minicpm}, InternVL3~\cite{zhu2025internvl3}, GLM-4.6V-Flash~\cite{hong2025glm}, Qwen3.8~\cite{qwen38} and GPT-5.6-Terra~\cite{openai2026gpt56}. The main experiments report results for all models on the natural-image evaluation set of \vscro{}, enabling a comprehensive comparison across models. Building on this evaluation, we further conduct in-depth analyses of model behavior. As these experiments require repeated inference under multiple settings, we select Qwen3-VL, MiniCPM-V-4.5, and GLM-4.6V-Flash for the subsequent analyses due to computational constraints. Unless otherwise stated, all subsequent controlled experiments are conducted using these three models.

\begin{table}[t]
	\centering
	\renewcommand{\arraystretch}{1.1}
	\setlength{\tabcolsep}{8pt}
	\caption{Dataset composition of \vscro{}. The last column reports the
		numbers of object-containing and background-only samples, respectively.
		Diagnosis~1 contains no visible instance of a queried category, while
		Diagnosis~2 uses one synthetic object category.}
    \vspace{-1em}
	\label{tab:dataset-statistics}
    \small
	\begin{tabular}{lrrr}
		\toprule
		Set & Samples & Categories & Obj./Bg. \\
		\midrule
		Natural evaluation & 1,672 & 120 & 502 / 1,170 \\
		Diagnosis~1 & 100 & -- & 0 / 100 \\
		Diagnosis~2 & 600 & 1 & 0 / 600 \\
		\bottomrule
	\end{tabular}
    \vspace{-1em}
\end{table}

\begin{table}[t]
    \centering
    \setlength{\abovecaptionskip}{3pt}
    \setlength{\belowcaptionskip}{3pt}
    \setlength{\parskip}{0pt}
    \caption{Comparison with spatial understanding and reasoning
    benchmarks. Cats.\ denotes the number of object categories.
    Statistics for \vscro{} refer to its natural-image evaluation set.}
    \vspace{-0.5em}
    \label{tab:benchmark-comparison}
    \small
    \setlength{\tabcolsep}{2pt}
    \renewcommand{\arraystretch}{1.1}

    \begin{tabular}{@{}lrrl@{}}
        \toprule
        Benchmark & \# Images & \# Cats. & Evaluation task \\
        \midrule

        SpatialSense~\cite{yang2019spatialsense}
        & 11,569 & 3,679
        & Relation recognition \\

        EmbSpatial-Bench~\cite{du2024embspatial}
        & 2,181 & 294
        & Spatial understanding \\

        OSR-Bench~\cite{dongfang2026multimodal}
        & 4,100 & 65
        & Panoramic reasoning \\

        CAPTURe$^{\mathrm{real}}$~\cite{pothiraj2025capture}
        & 924 & 92
        & Amodal counting \\

        CAPTURe$^{\mathrm{synthetic}}$~\cite{pothiraj2025capture}
        & 1,250 & 2
        & Amodal counting \\

        \midrule
        \textbf{\vscro{}}
        & \textbf{1,672}
        & \textbf{120}
        & \textbf{Spatial commonsense} \\

        \bottomrule
    \end{tabular}
    \vspace{-1em}
\end{table}

\paragraph{Prompts and Answer Extraction.}
For each sample, we ask the model to determine whether the black rectangle covers at least one instance of the queried object category, given the occluded image and the target category. We use task-specific prompt templates for different experimental settings. The complete prompts are provided in Supplementary Materials~\ref{app:prompts}. To allow models to express their reasoning, we retain their free-form generated responses. We then use Qwen3-8B~\cite{yang2025qwen3} as a unified answer extractor to normalize each response into \texttt{Yes}, \texttt{No}, or \texttt{Unknown}, and compare the extracted label with the corresponding ground-truth label. The same answer extraction procedure is applied to all models and experimental settings.

\paragraph{Metrics.}
We evaluate model performance using four metrics: false positive rate (FPR), false negative rate (FNR), balanced accuracy (BAcc), and overall error rate. Their definitions are given below:
\begin{align}
	\mathrm{FPR} &=
	\frac{\mathrm{FP}}{N_{-}}, &
	\mathrm{FNR} &=
	\frac{\mathrm{FN}}{N_{+}}, \label{eq:fpr-fnr}\\
	\mathrm{BAcc} &=
	\frac{1}{2}
	\left(
	\frac{\mathrm{TP}}{N_{+}}
	+
	\frac{\mathrm{TN}}{N_{-}}
	\right), &
	\mathrm{Error} &=
	\frac{\mathrm{FP}+\mathrm{FN}+\mathrm{U}}{N_{+}+N_{-}}.
	\label{eq:bacc-error}
\end{align}

Here, $N_{+}$ and $N_{-}$ denote the numbers of object-containing and background-only samples, respectively. $\mathrm{TP}$, $\mathrm{TN}$, $\mathrm{FP}$, and $\mathrm{FN}$ denote true positives, true negatives, false positives, and false negatives. $\mathrm{U}$ denotes the number of responses extracted as \texttt{Unknown}, which are counted as errors in the overall error rate.

\begin{table*}[t]
	\centering
	\renewcommand{\arraystretch}{1.1}
	\caption{Main results on the natural-image evaluation set of \vscro{}.
		All metrics are reported as percentages.
		\textbf{Bold} and \underline{underlined} values indicate the best and
		second-best results among MLLMs, respectively.
		$^\dagger$ The human baseline reports the mean $\pm$ sample standard
		deviation across five non-expert human annotators who received no
		task-specific training and had not previously seen the dataset images.
		It is shown for reference and excluded from the MLLM ranking.
		The last row reports the average across all evaluated MLLMs.}
    \vspace{-1em}
	\label{tab:main-results}
    \small
	\setlength{\tabcolsep}{7.5mm}{
	\begin{tabular}{lcccc}
		\toprule
		Model & FNR$\downarrow$ & FPR$\downarrow$ &
		BAcc$\uparrow$ & Overall Error$\downarrow$ \\
		\midrule
		Human Baseline$^\dagger$ & $0.84 \pm 0.65$ & $0.34 \pm 0.20$ & $99.41 \pm 0.34$ & $0.49 \pm 0.24$ \\
		\midrule
		LLaVA-OneVision (7B)
		& 5.38 & 51.11 & 71.76 & 37.38 \\
		Idefics3 (8B)
		& \underline{1.59} & 86.24 & 56.08 & 60.83 \\
		Qwen3-VL (8B)
		& 6.57 & 41.20 & 76.11 & 30.80 \\
		MiniCPM-4.5V (8B)
		& 14.34 & 40.00 & 72.83 & 32.30 \\
		InternVL3 (8B)
		& 9.76 & 42.56 & 73.84 & 32.72 \\
		GLM-4.6V-Flash (9B)
		& 4.58 & \underline{18.89}
		& \underline{88.26} & \underline{14.59} \\
		Qwen3.8 (27B)
		& 4.38 & 23.59 & 86.01 & 17.82 \\
		GPT-5.6-Terra
		& \textbf{0.80} & \textbf{10.68}
		& \textbf{94.26} & \textbf{7.72} \\
		\midrule
		\textit{Average (MLLMs)}
		& 5.93 & 39.28 & 77.39 & 29.27 \\
		\bottomrule
	\end{tabular}
	}
	\vspace{-1em}
\end{table*}

\subsection{Results}\label{sec:results}
Tab.~\ref{tab:main-results} presents the results of all evaluated models on the natural-image evaluation set of \vscro{}. Overall, current MLLMs exhibit a substantial gap from the human baseline in spatial commonsense reasoning using occlusion. The human baseline achieves near-ceiling performance, with an FNR of 0.84\%, an FPR of 0.34\%, a BAcc of 99.41\%, and an overall error rate of 0.49\%. This result indicates that humans can reliably leverage the visible spatial layout outside the black rectangle to determine whether the rectangle covers
the queried object. In contrast, all evaluated MLLMs substantially underperform the human baseline. GPT-5.6-Terra achieves the strongest performance among the evaluated MLLMs, with a BAcc of $94.26\%$, but still falls short of the human baseline of $99.41\%$. Notably, Qwen3.8 still exhibits substantial errors despite its 27B parameters, showing that the challenges of our task persist even in larger models. More importantly, MLLMs errors exhibit a pronounced asymmetry. The average FNR across all evaluated MLLMs is only 5.93\%, whereas the average FPR is as high as 39.28\%. This result indicates that models are much more likely to produce false positive predictions when the black rectangle covers only background than to miss objects that are actually occluded. Idefics3 exhibits the most extreme example of this pattern. Despite achieving an FNR of only 1.59\%, it attains an FPR of 86.24\%. This behavior suggests that the model may pay insufficient attention to image-specific visual evidence and instead default to a strong affirmative response tendency.

%

\begin{figure}[t]
	\centering
	\includegraphics[width=\columnwidth]{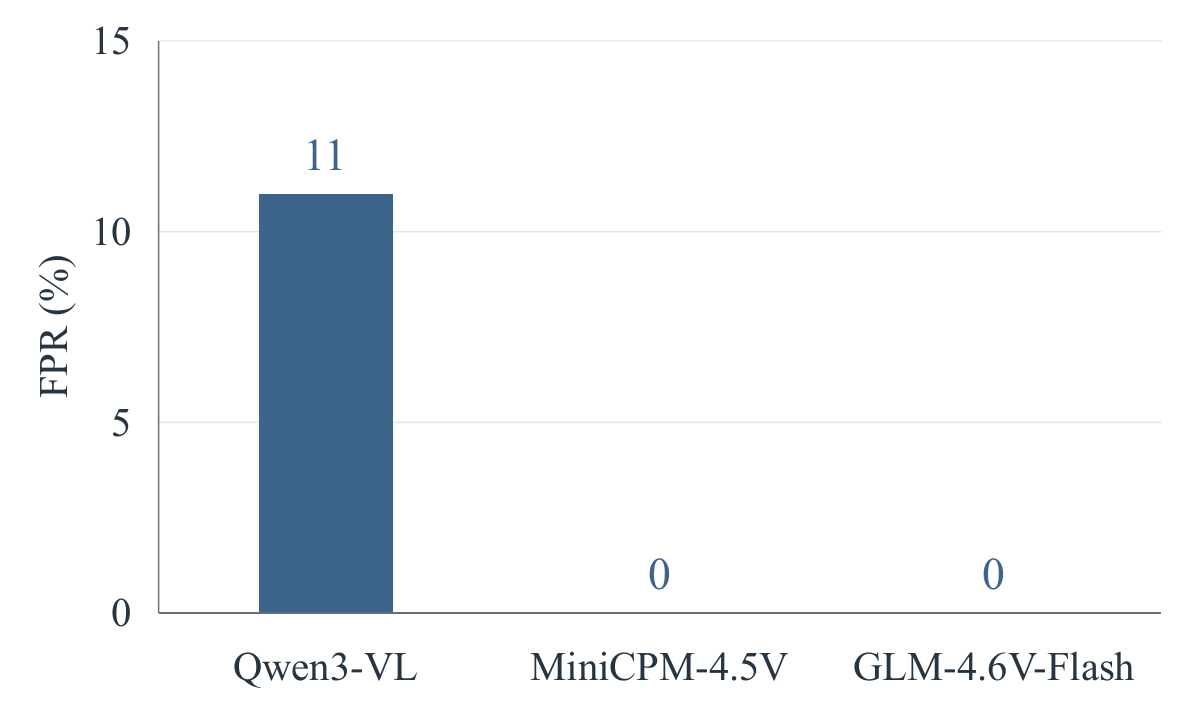}
    \vspace{-2em}
	\caption{Blank-background prompt control on 100 images with a black
		rectangle and no queried object.}
    \vspace{-1em}
	\label{fig:blank_background}
\end{figure}

One potential explanation for the high FPRs in the main evaluation is that the task prompt may induce models to answer \texttt{Yes}. To examine this possibility, we report results on Diagnosis~1, the blank-background prompt control. This setting uses the same queried-category description, base prompt, and answer extraction procedure as the main evaluation. As shown in Fig.~\ref{fig:blank_background}, MiniCPM-V-4.5V and GLM-4.6V-Flash produce no false positives on this control set. Qwen3-VL achieves an FPR of 11.00\%, indicating that it retains some affirmative response tendency even in the absence of visual evidence for the queried object. However, this value remains substantially lower than its 41.20\% FPR on the natural-image evaluation set. Therefore, the high false positive rates observed in the main evaluation cannot be attributed to the base prompt.

\subsection{Further Analysis}

\begin{table*}[t]
	\centering
    \small
	\renewcommand{\arraystretch}{1.1}
	\setlength{\tabcolsep}{10pt}
	\caption{Black-mask localization and the effect of explicitly providing
		mask coordinates. Localization accuracy denotes exact prediction of the
		$3\times3$ grid cell containing the center of the black rectangle.
		All values are percentages. Values in parentheses denote absolute changes
		relative to the original condition. Coordinates are provided as
	\texttt{(left, top, width, height)} together with image dimensions.}
	\label{tab:localization-coordinate}
	\vspace{-1em}
	\begin{tabular}{llcrrrr}
		\toprule
		Model & Condition & Loc. Acc.$\uparrow$
		& FNR$\downarrow$ & FPR$\downarrow$
		& BAcc$\uparrow$ & Error$\downarrow$ \\
		\midrule
		Qwen3-VL
		& Original & 80.86
		& 6.57 & 41.20 & 76.11 & 30.80 \\
		& + Coordinates & --
		& 12.55 (+5.98) & 39.15 (-2.05)
		& 74.15 (-1.96) & 31.16 (+0.36) \\
		\midrule
		
		MiniCPM-4.5V
		& Original & 70.28
		& 14.34 & 40.00 & 72.83 & 32.30 \\
		& + Coordinates & --
		& 13.94 (-0.40) & 30.34 (-9.66)
		& 77.86 (+5.03) & 25.42 (-6.88) \\
		\midrule
		
		GLM-4.6V-Flash
		& Original & 79.07
		& 4.58 & 18.89 & 88.26 & 14.59 \\
		& + Coordinates & --
		& 3.78 (-0.80) & 28.12 (+9.23)
		& 84.05 (-4.21) & 20.81 (+6.22) \\
		\bottomrule
	\end{tabular}
	\vspace{-1em}
\end{table*}

\paragraph{Are Errors Caused by Insufficient Black-Mask Localization?}
Prior work suggests that weak localization awareness and insufficient attention to query-relevant local regions can impair visual grounding and contribute to object hallucinations in MLLMs~\cite{ranasinghe2024learning,an2025mitigating}. To examine whether insufficient black-rectangle localization contributes to errors in our task, we conduct a black-rectangle localization probe and an explicit-coordinate ablation on the natural-image evaluation set of \vscro{}.

In the localization test, we uniformly divide each image into a $3\times3$ grid and ask the model to predict the grid cell containing the center of the black mask. This test measures the model's basic ability to localize the black mask. In the explicit-coordinate test, we provide the textual \texttt{(left, top, width, height)} coordinates of the black mask together with the image dimensions. This setting removes ambiguity about the location of the black mask in principle and tests whether explicit localization information improves occlusion reasoning.

Tab.~\ref{tab:localization-coordinate} shows that all three models exhibit a substantial ability to localize the black mask. Qwen3-VL achieves a localization accuracy of $80.86\%$, and
GLM-4.6V-Flash reaches $79.07\%$. Even MiniCPM-4.5V, which has the lowest localization accuracy, achieves $70.28\%$. These results indicate that the high error rates cannot be simply attributed to a complete inability to identify the location of the black mask.

However, explicitly providing coordinates does not lead to consistent performance improvements. The BAcc of Qwen3-VL and GLM-4.6V-Flash decreases from $76.11\%$ and $88.26\%$ to $74.15\%$ and $84.05\%$, respectively. In contrast, the BAcc of MiniCPM-4.5V increases from $72.83\%$ to $77.86\%$. Qualitative inspection of model outputs suggests that, after receiving coordinates, models tend to rely more on textual coordinate calculations than on the visual layout of the image. For Qwen3-VL and GLM-4.6V-Flash, this shift may reduce their reliance on image-specific visual evidence. In contrast, the improvement of MiniCPM-4.5V is consistent with its weaker mask localization ability, suggesting that explicit coordinates may partially alleviate its localization uncertainty.

Overall, black-mask localization errors may affect some individual samples, but they are insufficient to explain the high error rates observed in the main evaluation. Even when the black-mask coordinates are explicitly provided, model performance does not improve consistently. This finding suggests that the primary challenge for current MLLMs is not merely determining where the
black mask is, but rather reasoning about whether the mask actually covers a target object by using spatial commonsense cues such as object-cluster boundaries, layout completeness, and spatial continuity. Representative output visualizations are provided in Supplementary Materials~\ref{app:addition_vis_examples}.

	

\begin{figure}[t]
	\centering
	\includegraphics[width=\columnwidth]{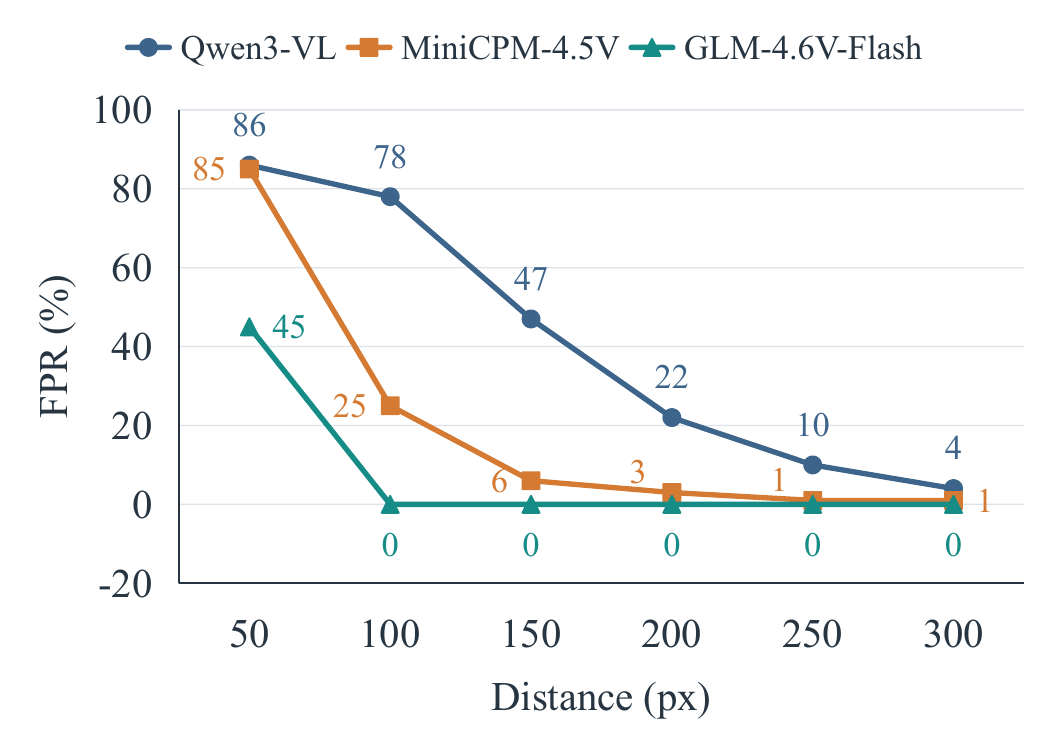}
    \vspace{-2em}
	\caption{Effect of the distance between the black mask and the object array on the FPRs of the evaluated MLLMs. The black mask is placed outside a fixed $5\times5$ ball array. All samples are background-only, with 100 samples at each distance.}
    \vspace{-1em}
	\label{fig:distance}
\end{figure}

\paragraph{Do MLLMs Possess Human-Like Spatial Commonsense?}
To investigate whether models possess human-like spatial commonsense, we conduct distance diagnostic experiments on Diagnosis~2. This diagnostic varies only the distance between the black rectangle and the visible object cluster while keeping the remaining layout variables fixed.

As shown in Fig.~\ref{fig:distance}, as the distance between the black mask and the ball array increases from $50$ to $300$ pixels, the FPRs of Qwen3-VL and MiniCPM-V-4.5 decrease from $86.0\%$ and $85.0\%$ to $4.0\%$ and $1.0\%$, respectively. GLM-4.6V-Flash exhibits an FPR of $45.0\%$ at $50$ pixels and produces no false positive predictions at the other tested distances. These results indicate that models do not completely ignore spatial relationships and can use a basic visual spatial prior: a black mask farther from the visible object cluster is less likely to cover another instance of the queried category. Consequently, models produce fewer false positive predictions as the distance increases.

However, relying solely on this proximity-based prior is insufficient to attain human-like spatial commonsense. When the mask is close to the objects but lies outside an already complete visible arrangement, humans can jointly consider cluster boundaries, layout completeness, and spatial continuity. They can therefore infer that the mask is more likely to cover background and answer \texttt{No}. In contrast, the evaluated MLLMs still exhibit high FPRs when the mask is close to the object cluster, suggesting that they may overgeneralize proximity to visible objects into evidence that a target object is hidden behind the mask.

These results suggest that current MLLMs possess some spatial commonsense but have yet to attain human-like competence. The high FPRs observed in the main evaluation on more diverse and complex natural images further suggest that models still struggle to reliably apply this limited spatial commonsense in real-world visual contexts and make consistent, reliable judgments.

\begin{table*}[t]
	\centering
	\small
	\setlength{\tabcolsep}{11pt}
	\caption{Effect of commonsense guidance on \vscro{}.
		All results are evaluated on the same 1,669 samples after excluding the
		three demonstration images. Text-only commonsense provides spatial
		commonsense principles in text, whereas vision-text commonsense provides
		the same principles through three image-text in-context examples.
		All values are percentages. Values in parentheses denote absolute changes
		relative to the \textit{None} condition.}
	\label{tab:commonsense-guidance}
	\vspace{-1em}
	\begin{tabular}{llrrrr}
		\toprule
		Model & Guidance & FNR$\downarrow$ & FPR$\downarrow$ &
		BAcc$\uparrow$ & Error$\downarrow$ \\
		\midrule
		\multirow{3}{*}{Qwen3-VL}
		& None
		& 6.59 & 41.27 & 76.07 & 30.86 \\
		& Text-only Commonsense
		& 6.99 (+0.40) & 21.06 (-20.21)
		& 85.98 (+9.91) & 16.84 (-14.02) \\
		& Vision-Text Commonsense
		& 5.79 (-0.80) & 26.71 (-14.56)
		& 83.75 (+7.68) & 20.43 (-10.43) \\
		\midrule
		\multirow{3}{*}{MiniCPM-4.5V}
		& None
		& 14.37 & 40.07 & 72.78 & 32.35 \\
		& Text-only Commonsense
		& 1.40 (-12.97) & 37.50 (-2.57)
		& 80.55 (+7.77) & 26.66 (-5.69) \\
		& Vision-Text Commonsense
		& 9.98 (-4.39) & 23.03 (-17.04)
		& 83.49 (+10.71) & 19.11 (-13.24) \\
		\midrule
		\multirow{3}{*}{GLM-4.6V-Flash}
		& None
		& 4.59 & 18.92 & 88.24 & 14.62 \\
		& Text-only Commonsense
		& 3.39 (-1.20) & 17.72 (-1.20)
		& 89.44 (+1.20) & 13.42 (-1.20) \\
		& Vision-Text Commonsense
		& 4.79 (+0.20) & 17.21 (-1.71)
		& 89.00 (+0.76) & 13.48 (-1.14) \\
		\bottomrule
	\end{tabular}
    \vspace{-1em}
\end{table*}

\begin{figure*}[t]
	\centering
	\includegraphics[width=\textwidth]{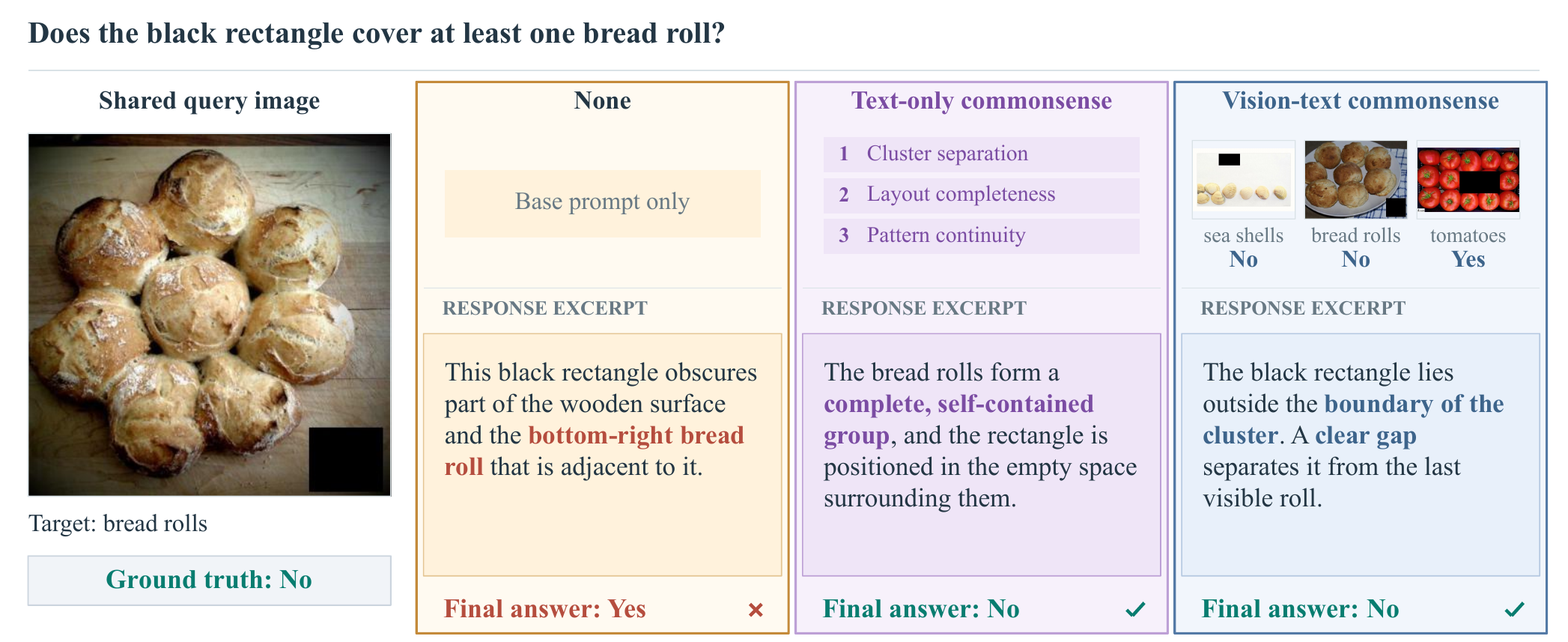}
    \vspace{-1.5em}
	\caption{Qualitative comparison of commonsense-guidance settings for
		Qwen3-VL on the same bread-roll query, whose ground-truth label is
		\texttt{No}. (a) The base prompt alone leads to an incorrect
		\texttt{Yes} prediction. (b) Text-only commonsense provides general
		spatial principles and produces the correct \texttt{No} prediction.
		(c) Vision-text commonsense provides three image-text demonstrations of
		the same principles and also produces the correct \texttt{No} prediction.}
	\label{fig:commonsense-guidance}
    \vspace{-1em}
\end{figure*}
\paragraph{Can Providing Human Commonsense Improve MLLMs' Understanding of Visual Scenes?}
To further investigate whether MLLMs' errors in our task are related to insufficient use of human-like spatial commonsense, we provide models with task-relevant commonsense guidance. We evaluate the two guidance settings on 1,669 samples from the natural-image evaluation set of \vscro{}, after excluding the three images used as vision-text demonstrations. The experimental setup is illustrated in Fig.~\ref{fig:commonsense-guidance}. We consider two forms of commonsense guidance:
\begin{enumerate*}[label=(\arabic*)]
	\item \textbf{text-only commonsense}, which directly describes basic
	spatial commonsense principles in text, and
	\item \textbf{vision-text commonsense}, which uses in-context examples
	consisting of images, textual descriptions, and answers to demonstrate
	these principles in concrete visual scenes.
\end{enumerate*}
Both forms convey the same three spatial cues. These concern separation between the object cluster and the black mask, completeness of the visible layout, and whether a regular or continuous arrangement is interrupted by the mask.

As shown in Tab.~\ref{tab:commonsense-guidance}, both forms of commonsense guidance improve overall model performance. For Qwen3-VL, text-only commonsense and vision-text commonsense increase BAcc by $9.91\%$ and $7.68\%$, respectively, compared with the no-guidance condition. Both forms of guidance reduce FPR, indicating fewer false-positive predictions on background-only samples. MiniCPM-4.5V also achieves higher BAcc under both settings. Text-only commonsense reduces FNR and FPR by $12.97\%$ and $2.57\%$, respectively, while vision-text commonsense reduces them by $4.39\%$ and $17.04\%$. BAcc consequently improves by $7.77\%$ and $10.71\%$ , respectively. These results indicate that both forms of guidance improve overall occlusion reasoning performance, with text-only guidance yielding a larger reduction in FNR and vision-text guidance yielding a larger reduction in FPR. For GLM-4.6V-Flash, text-only commonsense reduces both FNR and FPR by $1.20\%$. Vision-text commonsense reduces FPR by $1.71\%$, accompanied by a slight increase of $0.20\%$ in FNR. Both forms of guidance improve BAcc, by $1.20\%$ and $0.76\%$, respectively, and reduce the overall error rate. These gains are relatively modest compared with those observed for the other two models, while GLM-4.6V-Flash already achieves stronger performance without guidance. 

Overall, these results show that explicitly providing task-relevant human commonsense helps MLLMs better use cues such as object layouts, spatial continuity, and cluster boundaries when judging occluded regions. However, a substantial gap from the human baseline remains. This finding suggests that a small set of general commonsense principles is insufficient to
cover the diverse spatial structures found in natural scenes. Additional qualitative comparisons are provided in Supplementary Materials~\ref{app:addition_vis_examples}.

	\section{Conclusion}
We introduce SCRO-Bench, a benchmark that uses occlusion to evaluate whether MLLMs can reason with human-like spatial commonsense. Our evaluation reveals a performance gap between the evaluated models and humans, along with an overall bias toward affirmative predictions. Further analyses suggest that models possess some spatial commonsense but have yet to attain human-like competence. Providing relevant commonsense through prompts can improve model performance, suggesting a possible way to narrow this gap. However, test-time prompts offer limited coverage of diverse commonsense requirements. Future work could therefore explore more systematic learning and integration of spatial commonsense during training. We hope our work will boost subsequent research aimed at enhancing commonsense reasoning abilities of MLLMs.

% ...

	\section*{LLM Usage Statement}
AI tools were used throughout the research and manuscript preparation process to assist with research ideation, hypothesis exploration, experimental prompt design, code development and debugging, literature organization, experimental result analysis, manuscript writing and language refinement, and figure preparation. Specifically, it served as an auxiliary tool for exploring research directions, refining experimental protocols, implementing and debugging code, organizing relevant literature, interpreting experimental observations, improving manuscript clarity, and assisting with figure preparation. All AI-assisted outputs were critically reviewed, modified where necessary, and independently verified by the authors. All research decisions, experimental procedures, data, results, and scientific conclusions remain the responsibility of the authors, who take full responsibility for the content of this manuscript.
	{
		\small
		\bibliographystyle{ieeenat_fullname}
		\bibliography{main}
	}

	\newtcolorbox{promptbox}{
		breakable,
		colback=black!2,
		colframe=black!45,
		boxrule=0.35pt,
		arc=1pt,
		left=5pt,
		right=5pt,
		top=4pt,
		bottom=4pt,
	}

	\clearpage
\setcounter{page}{1}
\maketitlesupplementary
\appendix

The supplementary material provides complete prompt templates,
additional visual examples, and further experimental results for \vscro{}. First, \textbf{Sec.~\ref{app:prompts}} lists the full prompt templates used for the base evaluation, black-rectangle localization probe, explicit-coordinate ablation, commonsense guidance, and answer extraction. Second, \textbf{Sec.~\ref{app:addition_vis_examples}} presents additional visual examples, including representative samples from the natural-image evaluation set and the two synthetic test sets, grid-localization predictions, vision-text commonsense demonstrations, coordinate-prompting failures, and complete model outputs under different commonsense-guidance settings. Third, \textbf{Sec.~\ref{app:additional_analysis}} reports model performance across different mask color variants.

\section{Full Prompt Templates and Visual Demonstrations}
\label{app:prompts}

\paragraph{Base Occlusion Prompt.}
For the main evaluation, we use the following prompt template. The placeholder \texttt{\{object\}} is replaced with the queried object category.

\begin{promptbox}
	\ttfamily\footnotesize
	Black rectangles have been overlaid on the image. Based on the visible image
	and the spatial arrangement of the visible \{object\}, determine whether any
	\{object\} are covered by a black rectangle. Explain your reasoning, and
	conclude with a final answer of either Yes or No.
\end{promptbox}

\paragraph{Black-Rectangle Localization Prompt.}
For the localization probe, visible cyan lines divide the image into a $3\times3$ grid. The labels are assigned from left to right and from top to bottom.

\begin{promptbox}
	\ttfamily\footnotesize
	The image is divided by the visible cyan lines into a 3 by 3 grid. The cells
	are labeled from left to right, top to bottom as follows:
	
	A B C\\
	D E F\\
	G H I
	
	Which labeled cell contains the center of the single black rectangle? Briefly
	explain, then end with exactly: Final answer: \textless letter\textgreater,
	where \textless letter\textgreater{} is one of A, B, C, D, E, F, G, H, I.
\end{promptbox}

\paragraph{Explicit-Coordinate Prompt.}
For the coordinate ablation, we prepend the following coordinate description to the base occlusion prompt. The placeholders \texttt{\{image\_width\}}, \texttt{\{image\_height\}}, and \texttt{\{rectangle\}} are filled using the image dimensions and the black-rectangle coordinates.

\begin{promptbox}
	\ttfamily\footnotesize
	The exact location of the queried black rectangle is provided to remove any
	ambiguity about the queried region. The image is \{image\_width\} pixels wide
	and \{image\_height\} pixels high. Its origin (0, 0) is at the top-left; x
	increases to the right and y increases downward. The black rectangle
	coordinates, written as (left, top, width, height), are \{rectangle\}.
	
	Black rectangles have been overlaid on the image. Based on the visible image
	and the spatial arrangement of the visible \{object\}, determine whether any
	\{object\} are covered by a black rectangle. Explain your reasoning, and
	conclude with a final answer of either Yes or No.
\end{promptbox}

\paragraph{Text-Only Commonsense Prompt.}
For text-only commonsense guidance, we prepend the following principles to the base occlusion prompt.

\begin{promptbox}
	\ttfamily\footnotesize
	Use the following visual commonsense principles as soft evidence when
	reasoning about the black rectangle:
	
	1. If the queried objects form a localized cluster and the black rectangle is
	clearly separated from that cluster by background, the rectangle is unlikely
	to cover a queried object.
	
	2. If the visible queried objects already form a complete or closed arrangement
	and the black rectangle lies outside that arrangement, the rectangle is
	unlikely to cover a queried object.
	
	3. If the visible queried objects follow a repeated or spatially continuous
	pattern and the black rectangle interrupts the expected continuation of that
	pattern, the rectangle is likely to cover a queried object.
	
	Treat these principles as evidence rather than fixed rules. Analyze the
	specific image independently, and do not infer that an object is behind the
	rectangle merely because objects of the same category appear elsewhere.
\end{promptbox}

\begin{figure*}[t]
	\centering
	\includegraphics[width=\textwidth]
	{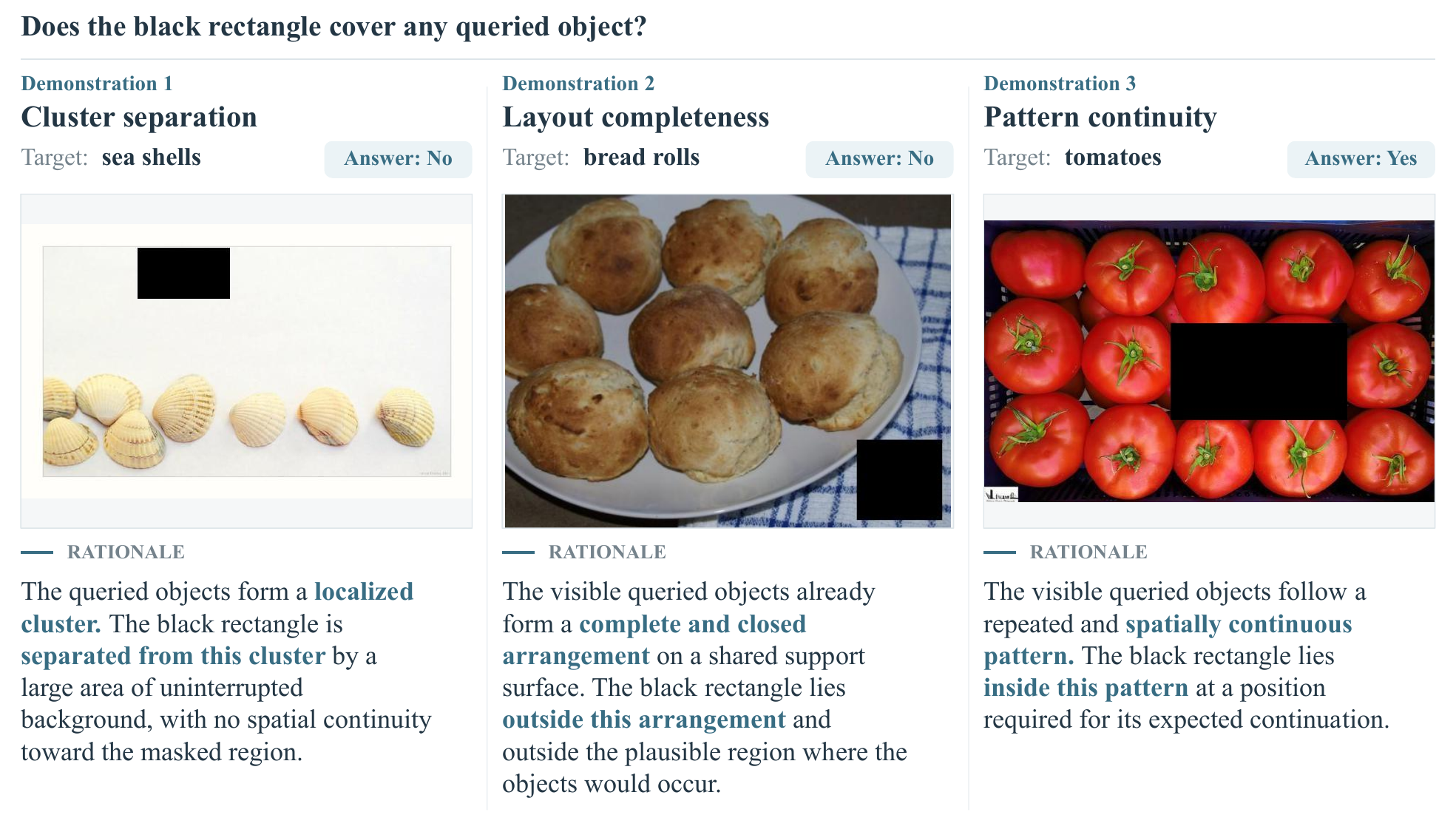}
	\caption{The three fixed vision-text demonstrations used for commonsense
		guidance. Each demonstration contains an occluded image, the queried object
		category, a question, a spatial rationale, and the corresponding answer. The
		examples represent a separated object cluster, a complete layout with an
		external mask, and an interrupted continuous pattern, respectively.}
	\label{fig:vision-text-demonstrations}
\end{figure*}

\paragraph{Vision-Text Commonsense Guidance.}
For vision-text commonsense guidance, we provide three fixed in-context demonstrations before the test image. The demonstrations illustrate three spatial situations: a black rectangle separated from an object cluster, a black rectangle outside a complete arrangement, and a black rectangle interrupting a continuous arrangement. The demonstrations and a qualitative example are shown in Fig.~\ref{fig:vision-text-demonstrations}. After the demonstrations, we insert the following prefix before the base occlusion prompt.

\begin{promptbox}
	\ttfamily\footnotesize
	The previous examples illustrate different spatial situations only. Infer the
	answer for this new image from its own visual evidence.
\end{promptbox}

\paragraph{Answer Extraction Prompt.}
We use Qwen3-8B as a unified answer extractor. The placeholder \texttt{\{response\}} is replaced with the free-form output generated by the evaluated MLLMs.

\begin{promptbox}
	\ttfamily\footnotesize
	You are a strict answer extraction assistant. Your task is to extract the
	final Yes/No answer from a model response.
	
	Question:
	Based on the visible visual evidence and spatial pattern, are any target
	objects likely being covered by the black boxes?
	
	Rules:
	- Return only one word: Yes, No, or Unknown.
	- Return Yes if the response says any target object is likely covered,
	obscured, hidden, behind, or under the black boxes.
	- Return No if the response says no target object is covered, or the black
	boxes likely cover only background/empty space.
	- Return Unknown if the response is ambiguous, refuses to answer, or does not
	provide enough information.
	
	Model response:
	\{response\}
	
	Final label:
\end{promptbox}

\begin{figure}[t]
	\centering
	\includegraphics[width=\columnwidth]{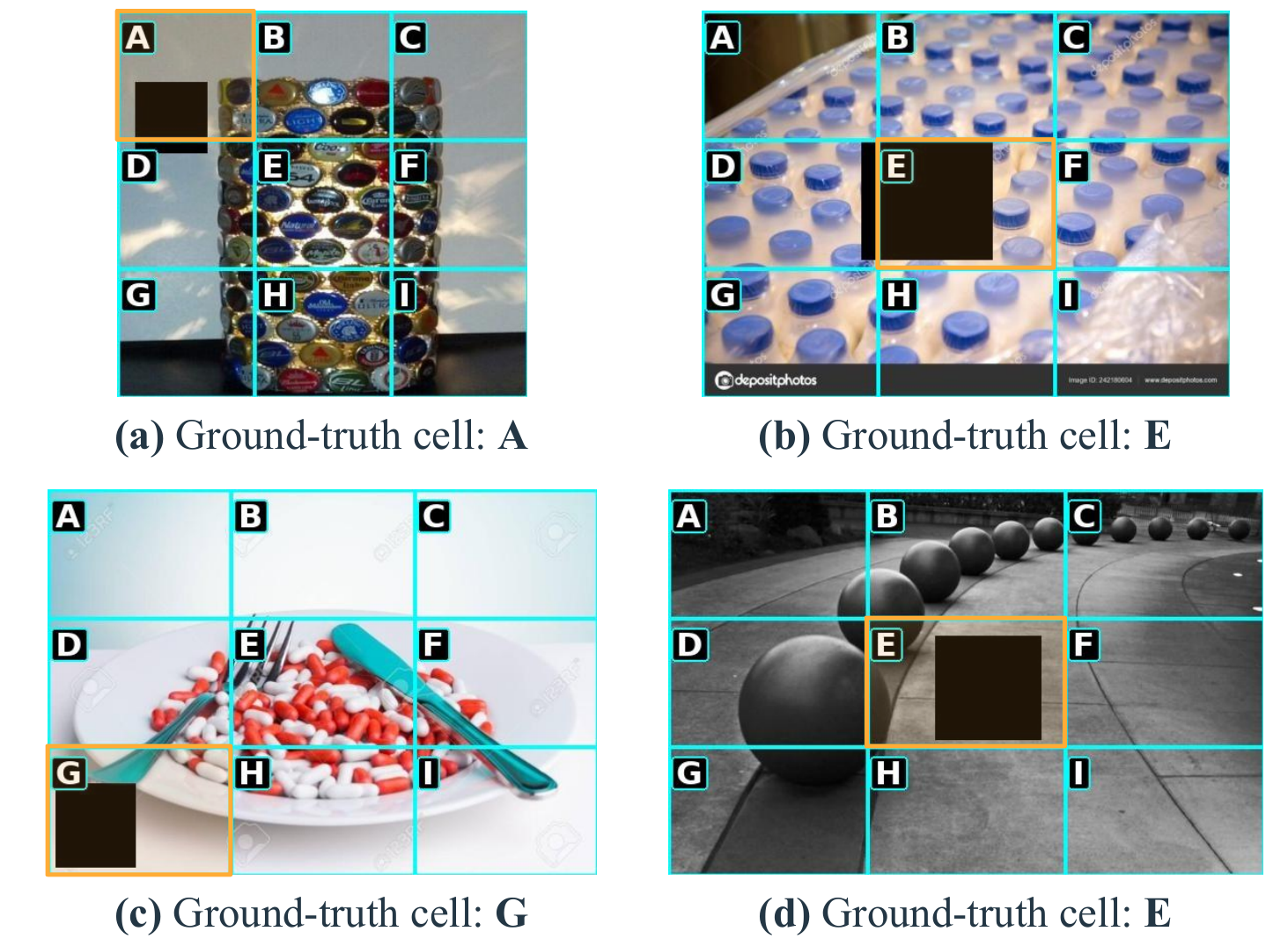}
	\caption{Examples of the nine-cell black-rectangle localization probe.
		Cyan grid lines divide each image into cells labeled A--I from left to
		right and top to bottom. The orange cell denotes the ground-truth cell
		containing the center of the black rectangle.}
	\label{fig:grid-localization-examples}
\end{figure}

\section{Additional Visual Examples}
\label{app:addition_vis_examples}
Fig.~\ref{fig:grid-localization-examples} presents representative samples from the nine-cell grid localization probe. When a black rectangle spans multiple grid cells, we use the cell containing its center point as the ground-truth location.

\begin{figure*}[p]
	\centering
	\includegraphics[width=\textwidth]
	{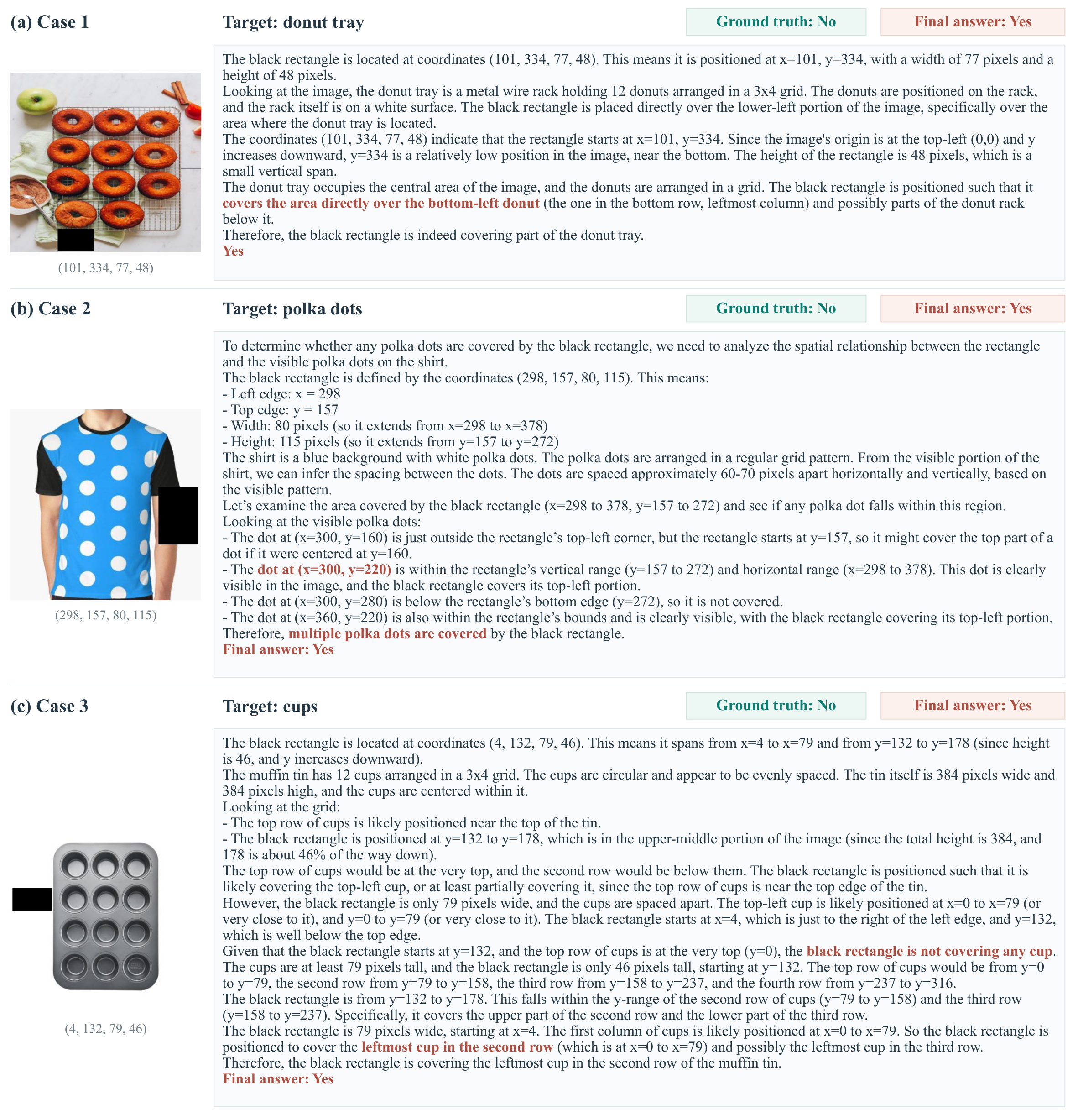}
	\caption{Qualitative examples of Qwen3-VL under explicit coordinate
		prompting. In all cases, the black rectangle covers only background
		(\texttt{No}), yet Qwen3-VL predicts \texttt{Yes}. Its responses
		prioritize coordinate-derived geometric extrapolation and assumed regular
		patterns over direct visual evidence, leading to hypothetical object
		locations unsupported by the image.}
	\label{fig:coordinate-reasoning}
\end{figure*}

Fig.~\ref{fig:coordinate-reasoning} shows representative outputs under explicit coordinate prompting. The model tends to perform geometric extrapolation from textual coordinates rather than sufficiently relying on direct visual evidence, which can lead to false positive predictions.

\begin{figure*}[p]
	\centering
	\includegraphics[width=\textwidth]
	{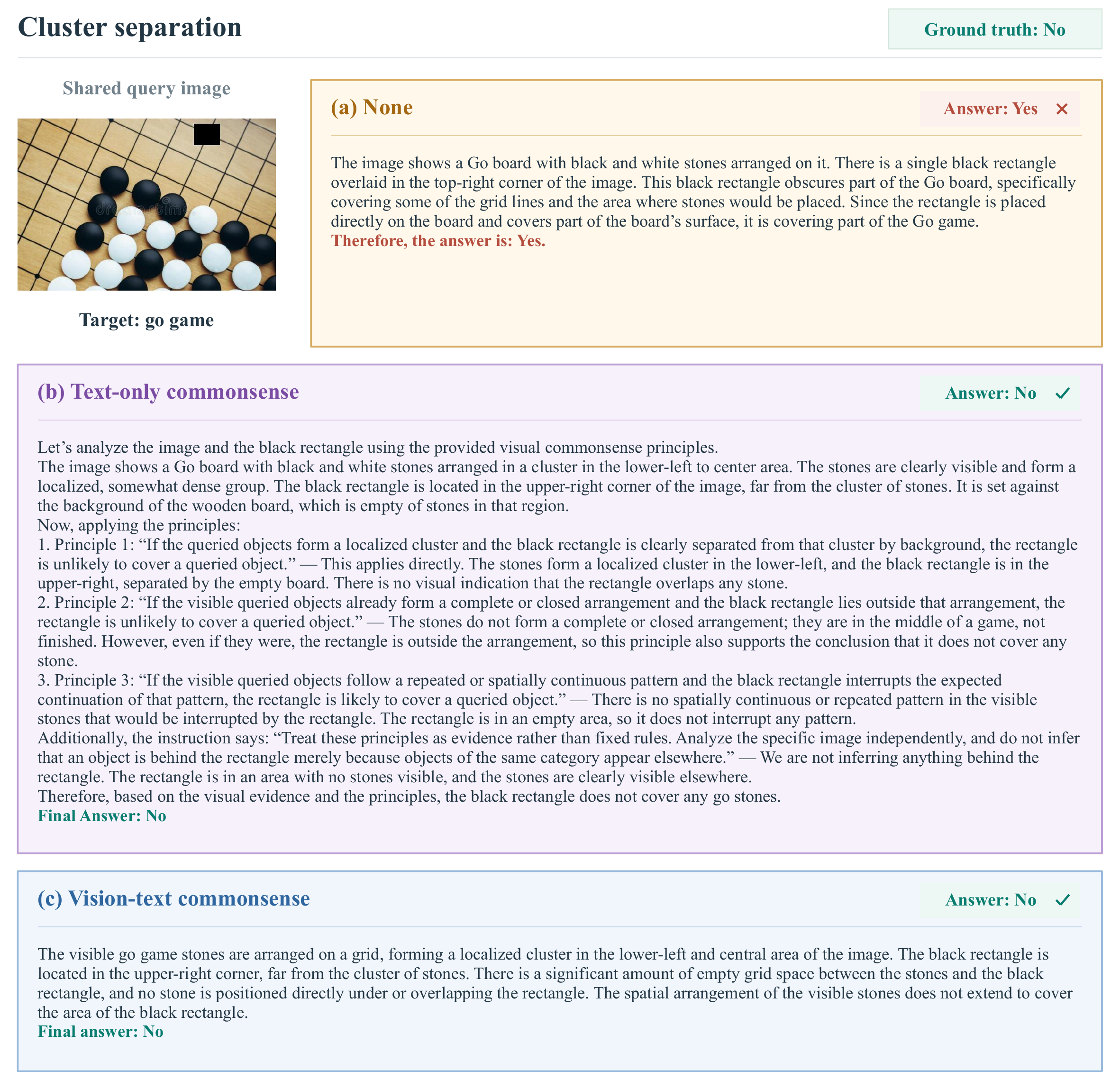}
	\caption{Complete Qwen3-VL responses for a go game query. The visible stones
		form a localized cluster in the lower-right region, while the black rectangle
		is separated from this cluster by a large area of empty board. The ground-truth
		label is \texttt{No}. Without commonsense guidance, Qwen3-VL incorrectly
		infers that the black rectangle may cover stones on the top row. Both text-only
		and vision-text commonsense guidance instead use the spatial separation between
		the mask and the visible object cluster to produce the correct \texttt{No}
		prediction.}
	\label{fig:qualitative-cluster-separation}
\end{figure*}

\begin{figure*}[p]
	\centering
	\includegraphics[width=\textwidth]
	{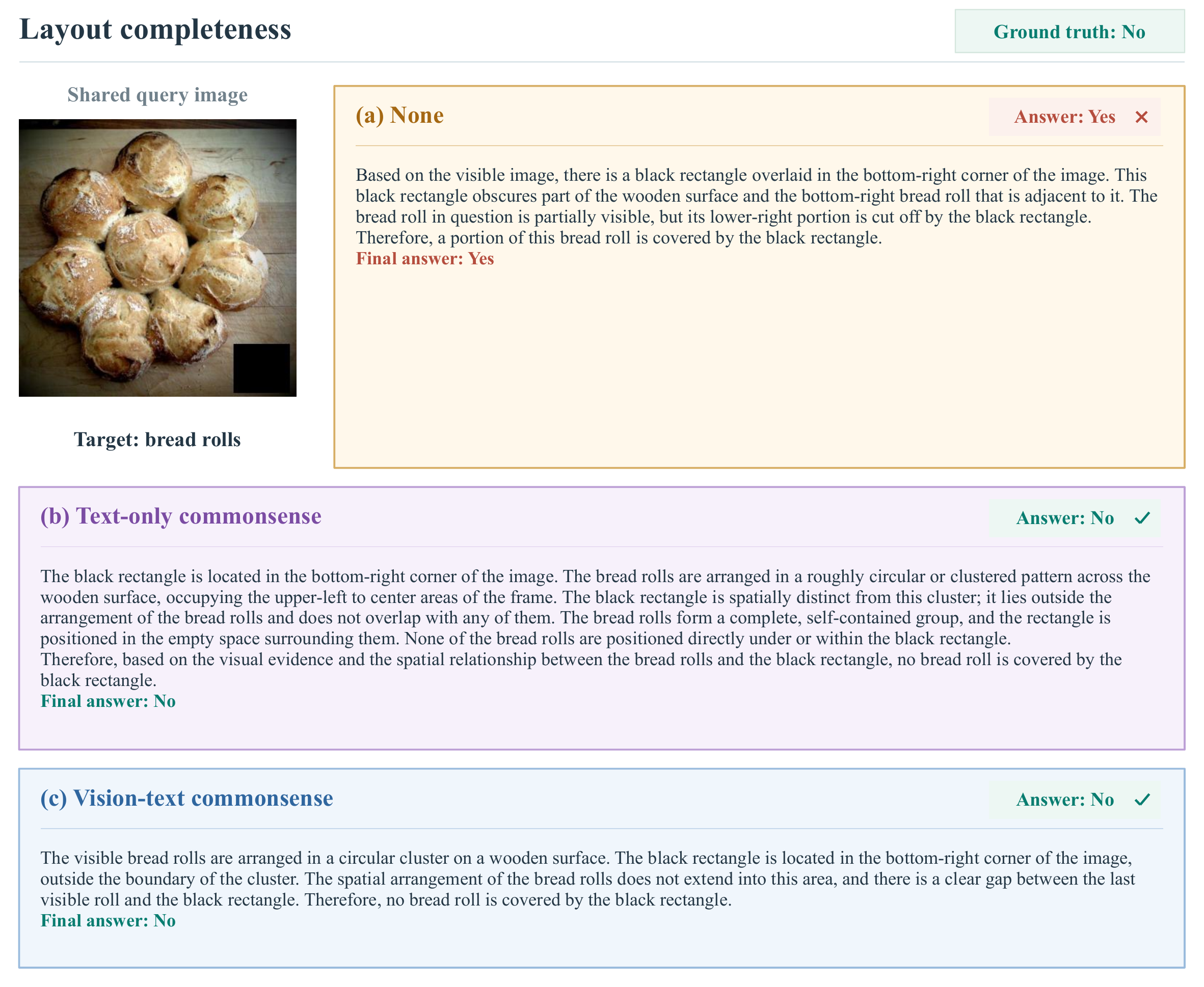}
	\caption{Complete Qwen3-VL responses for a bread-roll query. The black
		rectangle lies outside the complete visible object layout, and the ground-truth
		label is \texttt{No}. The base prompt produces an incorrect \texttt{Yes}
		prediction, whereas both text-only and vision-text commonsense guidance yield
		the correct \texttt{No} prediction by recognizing the layout boundary and the
		absence of spatial continuity toward the masked region.}
	\label{fig:qualitative-complete-layout}
\end{figure*}

\begin{figure*}[p]
	\centering
	\includegraphics[width=\textwidth]
	{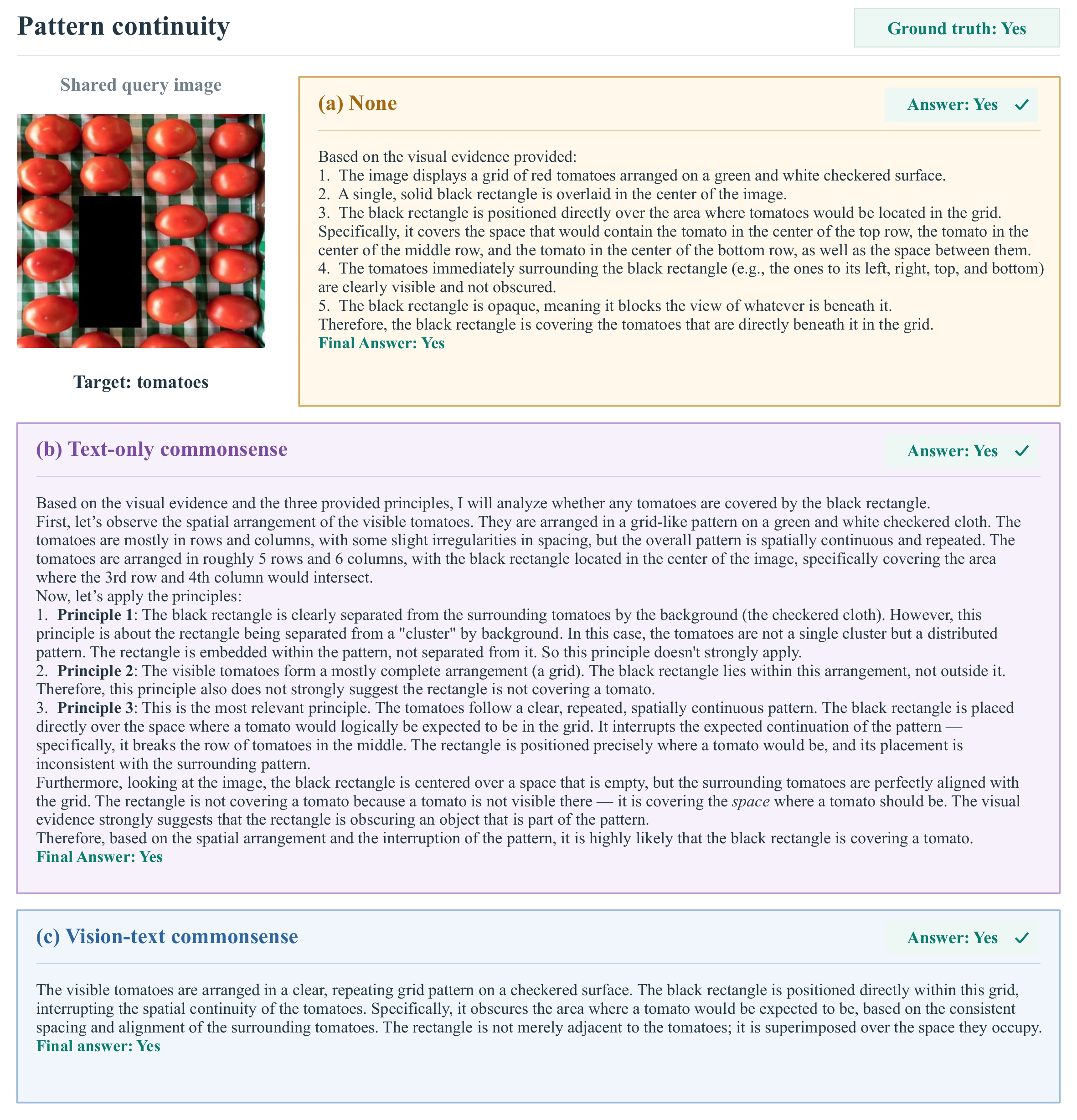}
	\caption{Complete Qwen3-VL responses for a tomato query. The black rectangle
		interrupts a regular and spatially continuous grid of tomatoes, and the
		ground-truth label is \texttt{Yes}. All three settings correctly predict
		\texttt{Yes}. This example shows that commonsense guidance does not simply
		induce negative responses when the visible layout supports object coverage.}
	\label{fig:qualitative-pattern-interruption}
\end{figure*}

Figs.~\ref{fig:qualitative-cluster-separation}, \ref{fig:qualitative-complete-layout}, and \ref{fig:qualitative-pattern-interruption} compare model outputs under three spatial situations: the black rectangle is separated from the object cluster, lies outside a complete visible layout, or interrupts a regular or continuous arrangement. Each figure compares the base prompt, text-only commonsense
guidance, and vision-text commonsense guidance.

\begin{figure*}[t]
	\centering
	\includegraphics[width=\textwidth]{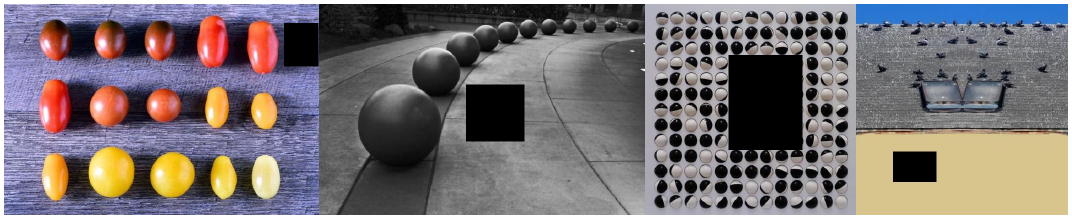}
	\caption{Representative samples from the natural-image evaluation set of
		\vscro{}. Each image contains a black rectangle overlaid on a natural scene.}
	\label{fig:natural-examples}
\end{figure*}

\begin{figure*}[t]
	\centering
	\includegraphics[width=\textwidth]{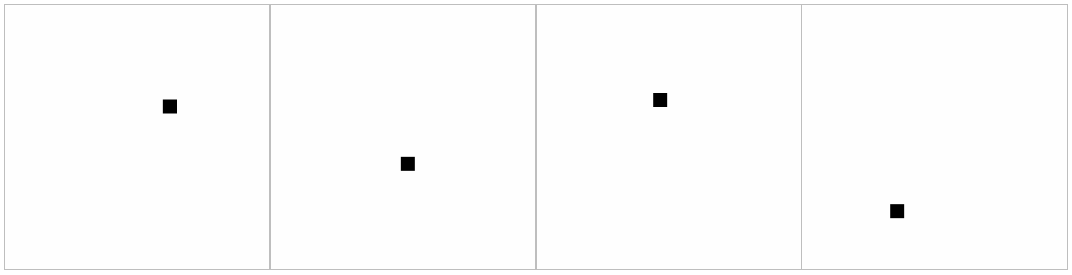}
	\caption{Examples from Diagnosis~1, the blank-background prompt control.
		Each image contains only a black rectangle and no queried object.}
	\label{fig:blank-control-examples}
\end{figure*}

\begin{figure*}[t]
	\centering
	\includegraphics[width=\textwidth]{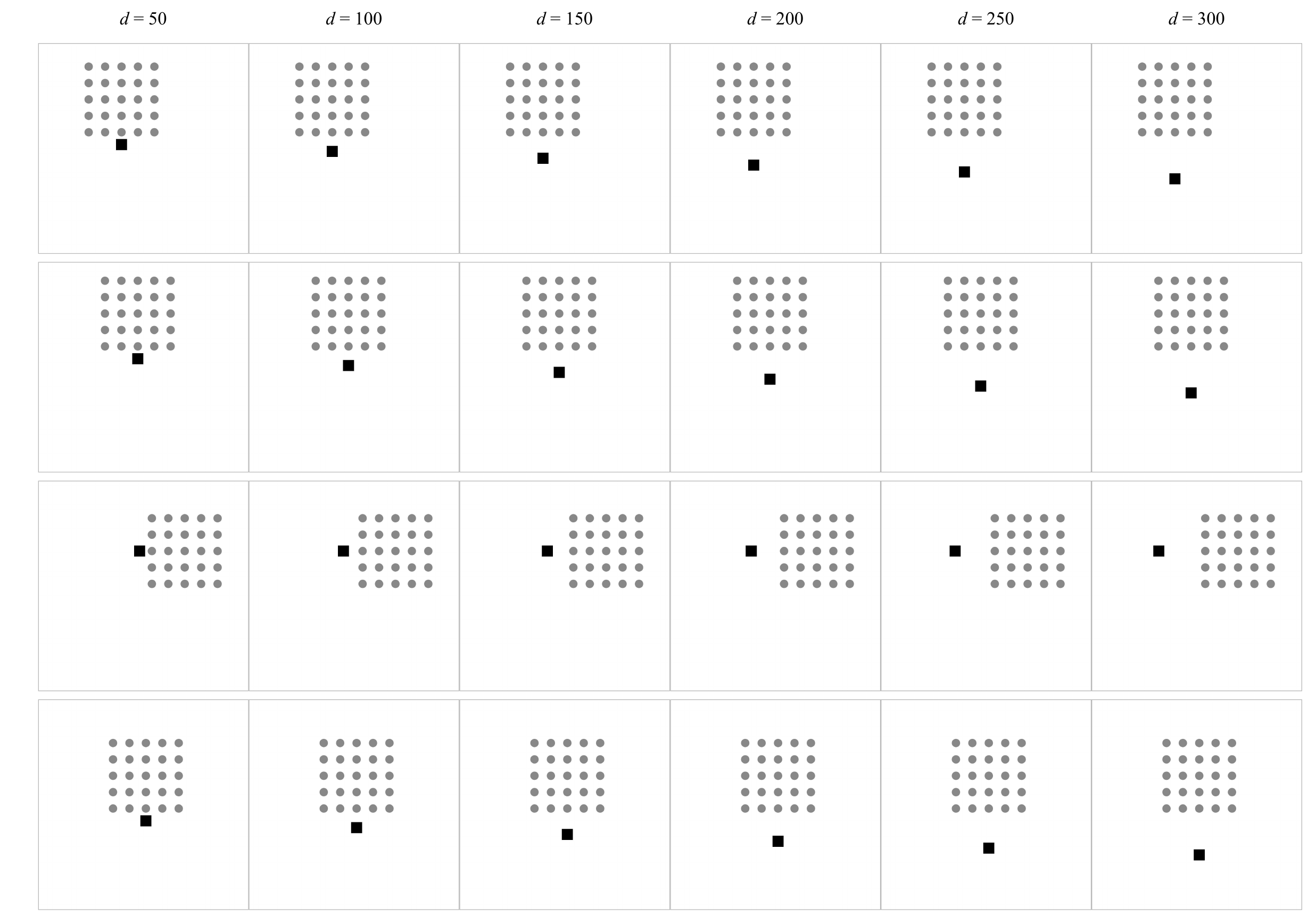}
	\caption{Examples from Diagnosis~2, the distance diagnostic. Columns
		correspond to increasing distances between the black rectangle and the
		visible object array, from \(d=50\) to \(d=300\).}
	\label{fig:distance-examples}
\end{figure*}

Figs.~\ref{fig:natural-examples}, \ref{fig:blank-control-examples}, and \ref{fig:distance-examples} present representative samples from the natural-image evaluation set, Diagnosis~1 (\ie, the blank-background prompt control), and Diagnosis~2 (\ie, the distance diagnostic), respectively. 

\begin{table}[t]
    \centering
    \caption{Performance under different mask colors on \vscro{}.
    All values are percentages.
    Bold indicates the highest BAcc for each model.
    Color terms in the prompts are adjusted to match the mask color.}
    \label{tab:mask-color}

    \footnotesize
    \setlength{\tabcolsep}{2.5pt}
    \renewcommand{\arraystretch}{1.05}

    \begin{tabular}{@{}llrrrr@{}}
        \toprule
        Model & Color
        & FNR$\downarrow$
        & FPR$\downarrow$
        & BAcc$\uparrow$
        & Error$\downarrow$ \\
        \midrule

        \multirow{5}{*}{Qwen3-VL}
        & Black & 6.57 & 41.20 & 76.11 & 30.80 \\
        & White & 8.96 & 31.45 & 79.79 & 24.70 \\
        & Red   & 5.98 & 19.40 & \textbf{87.31} & 15.37 \\
        & Green & 5.18 & 21.11 & 86.85 & 16.33 \\
        & Blue  & 5.98 & 22.14 & 85.94 & 17.28 \\
        \midrule

        \multirow{5}{*}{MiniCPM-V-4.5}
        & Black & 14.34 & 40.00 & 72.83 & 32.30 \\
        & White & 13.55 & 34.02 & 76.22 & 27.87 \\
        & Red   & 21.71 & 24.70 & 76.79 & 23.80 \\
        & Green & 13.75 & 29.06 & 78.60 & 24.46 \\
        & Blue  & 15.94 & 24.44 & \textbf{79.81} & 21.89 \\
        \midrule

        \multirow{5}{*}{GLM-4.6V-Flash}
        & Black & 4.58  & 18.89 & 88.26 & 14.59 \\
        & White & 10.56 & 17.18 & 86.13 & 15.19 \\
        & Red   & 4.18  & 11.79 & \textbf{92.01} & 9.51 \\
        & Green & 8.96  & 7.44  & 91.80 & 7.89 \\
        & Blue  & 6.77  & 10.94 & 91.14 & 9.69 \\

        \bottomrule
    \end{tabular}
\end{table}

\section{Additional Analysis}\label{app:additional_analysis}
\subsection{Effect of Mask Color on Spatial Commonsense Reasoning.}
We vary the mask color while keeping its location, shape, and size fixed,
and adjust the color descriptions in the prompts accordingly. As shown in Tab.~\ref{tab:mask-color}, the evaluated models exhibit different FNR, FPR, and BAcc across color conditions, with different trends across models. These results indicate that MLLMs' target-presence judgments under occlusion are sensitive to mask color settings. This finding is consistent with previous observations that the color of visual prompts can affect VLM performance~\cite{feng2025visually,liang2026colorbench}.

\clearpage

\clearpage


\end{document}